\documentclass{article} %
\usepackage{iclr2023_conference}

\usepackage[utf8]{inputenc} %
\usepackage[T1]{fontenc}    %
\usepackage{url}            %
\usepackage{nicefrac}       %
\usepackage{microtype}      %

\usepackage{amsmath,amsfonts,bm}

\def\Figref#1{Figure~\ref{#1}}

\def\eqref#1{equation~\ref{#1}}

\def\1{\bm{1}}

\DeclareMathAlphabet{\mathsfit}{\encodingdefault}{\sfdefault}{m}{sl}
\SetMathAlphabet{\mathsfit}{bold}{\encodingdefault}{\sfdefault}{bx}{n}

\newcommand{\R}{\mathbb{R}}

\newcommand{\softmax}{\mathrm{softmax}}

\DeclareMathOperator*{\argmax}{arg\,max}

\usepackage{graphicx}
\usepackage{booktabs} %
\usepackage{lipsum,booktabs}
\usepackage{algpseudocode}
\usepackage{amsfonts}
\usepackage{xcolor}
\usepackage[ruled,vlined]{algorithm2e}
\usepackage{hyperref}
\usepackage{enumitem}
\usepackage{amsmath}
\usepackage{amsbsy}
\usepackage{array,ragged2e}
\usepackage{lipsum}  
\usepackage{wrapfig}

\usepackage{multirow} %
\usepackage{soul}%
\usepackage{changepage,threeparttable} %

\newcolumntype{P}[1]{>{\RaggedRight\arraybackslash}p{#1}}

\title{Composing Task Knowledge with\\Modular Successor Feature Approximators}

\author{%
Wilka Carvalho\thanks{Contact author: \href{wcarvalh@umich.edu}{wcarvalh@umich.edu}.}\,\,$^{,1}$ \hspace{1em}
Angelos Filos$^{2}$          \hspace{1em} \\ \bf
Richard L. Lewis$^{1}$          \hspace{1em}
Honglak lee$^{1, 3}$          \hspace{1em} 
Satinder Singh$^1$          \hspace{1em}
\\[2pt]
$^1$University of Michigan \hspace{2em}
$^2$University of Oxford \hspace{2em}
$^3$LG AI Research \hspace{2em}
}

\newcommand*{\msfapath}{.}%

\iclrfinalcopy %
\begin{document}

\maketitle

\begin{abstract}
  Recently, the Successor Features and Generalized Policy Improvement (SF\&GPI) framework has been proposed as a method for learning, composing, and transferring predictive knowledge and behavior. SF\&GPI works by having an agent learn predictive representations (SFs) that can be combined for transfer to new tasks with GPI. However, to be effective this approach requires state features that are useful to predict, and these state-features are typically hand-designed. In this work, we present a novel neural network architecture, “Modular Successor Feature Approximators” (MSFA), where modules both discover what is useful to predict, and learn their own predictive representations. We show that MSFA is able to better generalize compared to baseline architectures for learning SFs and modular architectures for learning state representations.
\end{abstract}

\section{Introduction}

Consider a household robot that needs to learn tasks including picking up dirty dishes and cleaning up spills. Now consider that the robot is deployed and encounters a table with both a spill and a set of dirty dishes. Ideally this robot can combine its training behaviors to both clean up the spill and pickup the dirty dishes. We study this aspect of generalization: combining knowledge from multiple tasks.

Combining knowledge from multiple tasks is challenging because it is not clear how to synthesize either the behavioral policies or the value functions learned during training. 
This challenge is exacerbated when an agent also needs to generalize to novel appearances and environment configurations. 
Returning to our example, our robot might need to additionally generalize to both novel dirty dishes and to novel arrangements of chairs.

\begin{figure}[!htb]
  \centering
  \includegraphics[width=.7\textwidth]{\msfapath/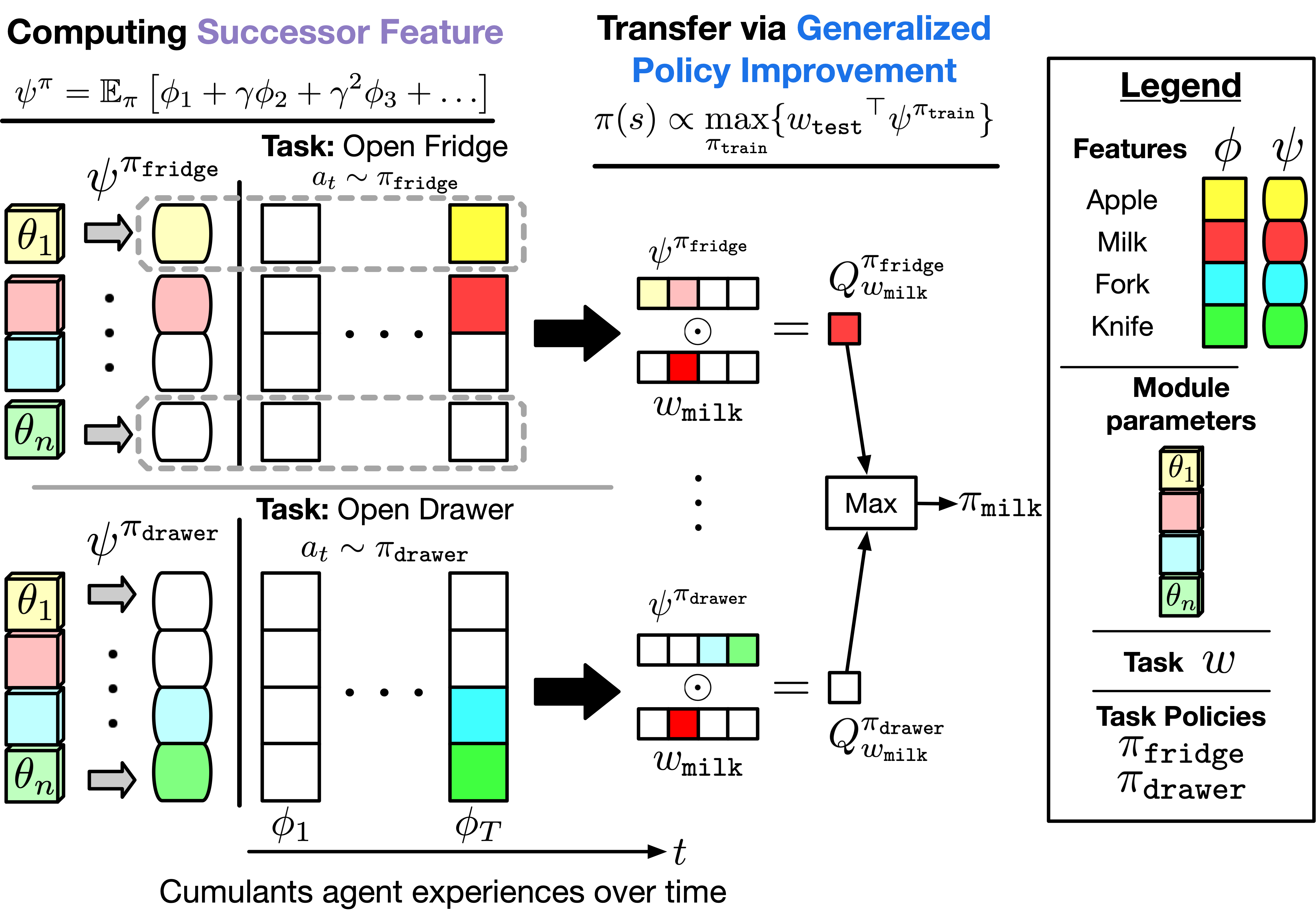}
  \caption{\textbf{High-level diagram of how MSFA can be leveraged for transfer with SF\&GPI}. During training, we can have the agent learn policies for tasks---e.g. ``open drawer'' and ``open fridge''. Each task leads the agent to experience different aspects of the environment---e.g. a ``fork'' during ``open drawer'' or an ``apple'' during ``open fridge''.
  We can leverage MSFA to have different modules learn different ``cumulants'', $\phi$, and SFs, $\psi$. For example, module 1 ($\theta_1$) can estimate SFs for apple cumulants. Module SFs are combined to form the SF for a policy. When the agent wants to transfer its knowledge to a test task---e.g., ``get milk''---it can compute Q-values for that task as a dot-product with the SFs of each training task. The highest Q-value is then used to select actions.  
    }\label{fig:high_level}
\end{figure}

Successor features (SFs) and Generalized Policy Improvement (GPI) provide a mechanism to combine knowledge from multiple training tasks~\citep{barreto2017successor,barreto2020fast}. SFs are predictive representations that estimate how much state-features (known as ``cumulants'') will be experienced given a behavior. By assuming that reward has a linear relationship between cumulants and a task vector, an agent can {efficiently} \textit{compute} how much reward it can expect to obtain from a given behavior. 
If the agent knows multiple behaviors, it can leverage GPI to compute which behavior would provide the most reward (see \Figref{fig:high_level} for an example). 
However, SF\&GPI commonly assume hand-designed cumulants and don't have a mechanism for generalizing to novel environment configurations.

Modular architectures are a promising method for generalizing to distributions outside of the training distribution~\citep{goyal2019recurrent,madan2021fast}. Recently, 
\citet{carvalho2021feature} presented ``FARM'' and showed that learning multiple state modules enabled generalization to environments with unseen environment parameters (e.g. to larger maps with more objects). 
In this work, we hypothesize that modules can further be leveraged to discover state-features that are useful to predict.

\begin{figure}[!htb]
\centering
\includegraphics[width=.7\textwidth]{\msfapath/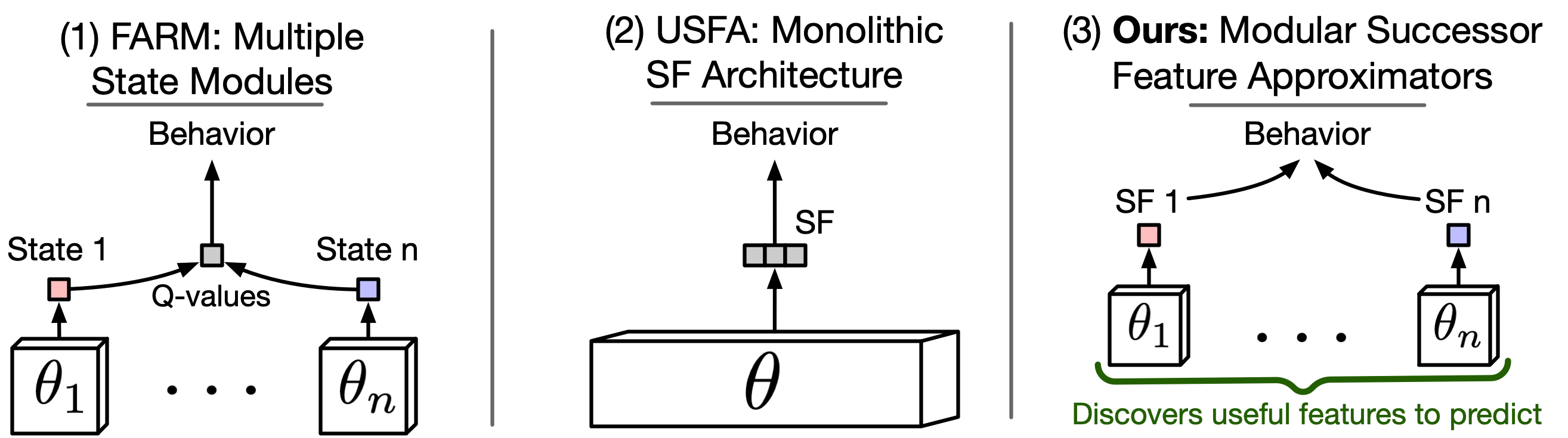}
\caption{ (1) FARM learns multiple state modules. This promotes generalization to novel environments. However, it has no mechanism for combining task solutions.
(2) USFA learns a single monolithic architecture for predicing SFs and can combine task solutions. However, it relies on hand-designed state features and has no mechanism for generalization to novel environments. 
(3) We combine the benefits of both. We leverage modules for reward-driven discovery of state features that are useful to predict. These form the basis of their own predictive representations (SFs) and enables combining task solutions in novel environments.}\label{fig:abstract_diagram}
\end{figure}
We present ``Modular Successor Feature Approximators'' (MSFA), a novel neural network for discovering, composing, and transferring predictive knowledge and behavior via SF\&GPI.
MSFA is composed of a set of modules, which each learn their own state-features and corresponding predictive representations (SFs).
\textbf{Our core contribution} is showing that an inductive bias for modularity can enable reward-driven discovery of state-features that are useful for zero-shot transfer with SF\&GPI. We exemplify this with a simple state-feature discovery method presented in \citet{barreto2018transfer} where the dot-product between state-features and a task vector is regressed to environment reward. This method enabled transfer with SF\&GPI in a continual learning setting but had limited success in the zero-shot transfer settings we study. While there are other methods for state-feature discovery, they add training complexity with mutual information objectives~\citep{hansen2019fast} or meta-gradients~\citep{veeriah2019discovery}. With MSFA, by adding \textit{only an architectural bias for modularity}, we discover state-features that (1) support zero-shot transfer competitive with hand-designed features, and (2) enable zero-shot transfer in visually diverse, procedurally generated environments. We are hopeful that our architectural bias can be leveraged with other discovery methods in future work.

\section{Related Work on Generalization in RL}\label{sec:related}

\textbf{Hierarchical RL} (HRL) is one dominant approach for combining task knowledge. The basic idea is that one can sequentially combine policies in time by having a ``meta-policy'' that sequentially activates ``low-level'' policies for protracted periods of time. 
By leveraging hand-designed or pre-trained low-level policies, one can generalize 
to longer instructions~\citep{oh2017zero,corona2020modular},
to new instruction orders~\citep{brooks2021reinforcement},
and to novel subtask graphs~\citep{sohn2020meta,sohn2022fast}.
We differ in that we focus on combining policies \textit{concurrently} in time as opposed to sequentially in time. To do so, we develop a modular neural network for the SF\&GPI framework.

\textbf{SFs} are predictive representations that represent the current state as a summary of the \textit{successive} features to follow (see \S\ref{sec:background} for a formal definition). By combining them with Generalized Policy Improvement, researchers have shown that they can transfer behaviors across object navigation tasks~\citep{borsa2018universal,zhang2017deep,zhu2017visual}, across continuous control tasks~\citep{hunt2019composing}, and within an HRL framework~\citep{barreto2019option}. However, these works tend to require hand-designed cumulants which are cumbersome to design for every new environment. In our work, we integrate SFs with Modular RL to facilitate reward-driven discovery of cumulants and improve successor feature learning.

\textbf{Modular RL} (MRL)~\citep{russell2003q} is a framework for generalization by combining value functions. Early work dates back to~\citep{singh1992transfer}, who had a mixture-of-experts system select between separately trained value functions. Since then, MRL has been applied to generalize
across robotic morphologies~\citep{huang2020one},
to novel task-robot combinations~\citep{devin2017learning,haarnoja2018composable},
and to novel language instructions~\citep{logeswaran2021learning}.
MSFA, is the first to integrate MRL with SF\&GPI. This integration enables combining task solutions in novel environment configurations.

\textbf{Generalizing to novel environment configurations with modules}.
\citet{goyal2019recurrent} showed that leveraging modules to learn a \textit{state function} improved out-of-distribution generalization.
\citet{carvalho2021feature} showed that a modified attention mechanism led to strong generalization improvements with RL.
MSFA differs from both in that it employ modules for learning \textit{value functions} in the form of SFs. This enables a principled way to compose task knowledge while additionally generalizing to novel environment configurations.

\section{Problem Setting and Background}\label{sec:background}
We study a reinforcement learning agent's ability to transfer knowledge between tasks in an environment.
During training, the experiences $n_{\tt train}$ tasks $\mathbb{M}_{\tt train} = \{\mathcal{M}_i\}_{i=1}^{n_{\tt train}}$, sampled from a training distribution $p_{\tt train}(\mathcal{M})$.
During testing, the agent is evaluated on $n_{\tt test}$ tasks, $\{\mathcal{M}_i\}_{i=1}^{n_{\tt test}}$, sampled from a testing distribution $p_{\tt test}(\mathcal{M})$. Each task $\mathcal{M}_i$ is specified as Partially Observable Markov Decision Process (POMDP), $\mathcal{M}_i = \langle 
\mathcal{S}^e, \mathcal{A}, \mathcal{X},
R, p, f_x
\rangle$. 
Here, $\mathcal{S}^e$, $\mathcal{A}$ and $\mathcal{X}$ are the environment state, action, and observation spaces. 
$p(\cdot|s^e_{t}, a_{t})$ specifies the next-state distribution based on taking action $a_{t}$ in state $s^e_{t}$, and $f_x(s^e_{t})$ maps the underlying environment state to an observation $x_t$. 
We focus on tasks where rewards are parameterized by a task vector $w$, i.e. $r^w_t = R(s^e_t, a_t, s^e_{t+1}, w)$ is the reward obtained for transition $(s^e_{t},a_{t},s^e_{t+1})$ given task vector $w$.
Since this is a POMDP, we need to learn a state function that maps histories to agent state representations. We do so with a recurrent function: $s_t = s_\theta(x_t, s_{t-1}, a_{t-1})$.
Given this learned state, we want to obtain a behavioral policy $\pi(s_t^{})$ that best maximises the expected reward it will obtain when taking an action $a_t$ at a state $s^{}_t$: $Q^{\pi, w}_t = Q^{\pi, w}(s^{}_t, a_t) = \mathbb{E}_{\pi}\left[\sum^{\infty}_{t=0}\gamma^t r^w_{t}\right]$.

\textbf{Transfer with SF\&GPI.} In order to leverage SFs~\citep{barreto2017successor}, one assumes an agent has access to state features known as ``cumulants'', $\phi_t = \phi(s^{}_t, a_t,s^{}_{t+1})$. Given a behavioral policy $\pi(a|s^{})$, SFs are a type of value function that use $\phi_t$ as pseudo-rewards:
\begin{equation}
  \psi^{\pi}_t 
  = \psi^{\pi}(s^{}_t, a_t) 
  = \mathbb{E}_{\pi} \left[\sum_{i=0}^{\infty} \gamma^i \phi_{t+i} \right]
\end{equation}
If reward is (approximately) $r^w_t = \phi_t^{\top}w$, then action-values can be decomposed as $Q^{\pi,w}_t = {\psi^{\pi}_t}^{\top}w$.
This is interesting because it provides an easy way to \textit{reuse} task-agnostic features $\psi^{\pi}_t$ for new tasks.

We can re-use the SFs we've learned from training tasks $\mathbb{M}_{\tt train}$ for transfer with GPI. 
Assume we have learned (potentially optimal) policies $\{\pi_i\}^{n_{\tt train}}_{i=1}$ and their corresponding SFs $\{\psi^{\pi_i}(s^{},a)\}^{n_{\tt train}}_{i=1}$.
Given a test task $w_{\tt test}$, we can obtain a new policy with GPI in two steps: (1) compute Q-values using the training task SFs (2) select actions using the highest Q-value. This operation is summarized as follows:
\begin{equation}
  \pi(s^{}_t;w_{\tt test}) \in \argmax_{a \in \mathcal{A}} \max_{i \in \{1, \ldots, n_{\tt train}\}} \{ {Q^{\pi_i, w_{\tt test}}_t}\}
   = \argmax_{a \in \mathcal{A}} \max_{i \in \{1, \ldots, n_{\tt train}\}} \{ {\psi^{\pi_i}_t}^{\top}w_{\tt test}\}
\end{equation}
This is useful because the GPI theorem states that $\pi$ will perform as well as all of the training policies, i.e. that $Q^{\pi, w_{\tt test}}(s^{},a) \geq \max_i Q^{\pi_i, w_i}(s^{},a) \forall (s^{},a) \in (\mathcal{S}^{} \times \mathcal{A})$~\citep{barreto2017successor}.

SF\&GPI enable transfer by exploiting structure in the RL problem: a policy that maximizes a value function is guaranteed to perform at least as well as the policy that defined that value function. However, SF\&GPI relies on combining a fixed set of SFs. Another form of transfer comes from ``Universal Value Function Approximators'' (UVFAs)~\citep{schaul2015universal}, which add the task-vector $w$ as a parameter to a Q-approximator parameterized by $\theta$, $Q_{\theta}(s^{},a,w)$. If $Q_{\theta}$ is smooth with respect to $w$, then $Q_{\theta}$ should generalize to test tasks nearby to train tasks in task space.~\citet{borsa2018universal} showed that one could combine the benefits of both with ``Universal Successor Feature Approximators''. Since rewards $r^w$, and therefore task vectors $w$, reference deterministic task policies $\pi_w$, one can parameterize successor feature approximators with task-vectors $\tilde{\psi}^{\pi_w} = \tilde{\psi}^{w} \approx \psi_{\theta}(s^{},a,w)$. 
However, USFA assumed hand-designed cumulants. We introduce an architecture for reward-driven discovery of cumulants and improved function approximation of universal successor features.

\begin{figure}[!htb]
  \centering
  \includegraphics[width=.7\textwidth]{\msfapath/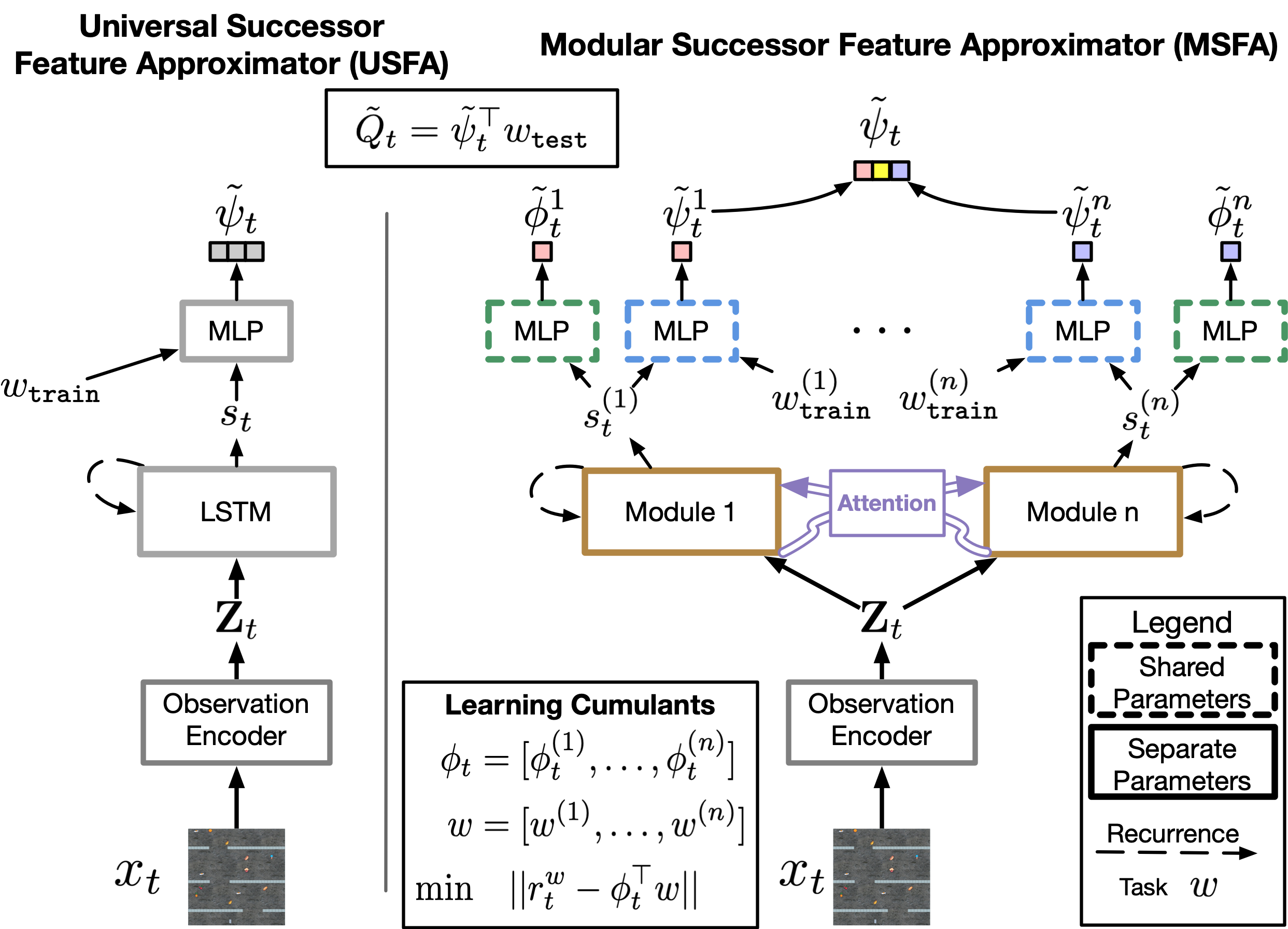}
  \caption{
    \textbf{Left: Universal Successor Feature Approximator (USFA)} learns a single, monolithic successor feature estimator that uses \textbf{hand-designed} cumulants.
    \textbf{Right: Modular Successor Feature Approximator (MSFA)} learns a set of successor feature modules, each with their own functions for (a) updating module-state, (b) computing cumulants, and (c) estimating successor features. Modules then share information with an attention mechanism. We hypothesize that isolated module computations facilitate learning cumulants that suppot generalization with GPI.
    }\label{fig:arch}
\end{figure}

\section{Modular Successor Feature Approximators}
We propose a new architecture \textit{Modular Successor Feature Approximators} (MSFA) for approximating SFs, shown in \Figref{fig:arch}.
Our hypothesis is that learning cumulants and SFs with modules improves zero-shot composition of task knowledge with SF\&GPI.
MSFA accomplishes this by learning $n$ state modules $\{s_{\theta_k}\}_{k=1}^n$ that evolve with independent parameters $\theta_k$ and have sparse inter-module information flow.
MSFA then produces modular cumulants $\{\tilde{\phi}^{(k)}_t\}^n_{k=1}$ and SFs $\{\tilde{\psi}^{\pi,k}_t\}^n_{k=1}$ by having their computations depend \textbf{only} on individual module-states.
For example, a cumulant may correspond to information about apples, and would be a function \textbf{only} of the module representing state information related to apples.
This is in contrast to prior work, which learns a single monolithic prediction module for computing cumulants and SFs (see \Figref{fig:arch}).

The rest of section is structured as follows.
In section \S\ref{sec:msfl}, we derive Modular Successor Feature Learning within the Modular RL framework. We then describe our architecture, MSFA, for learning modular successor features in \S\ref{sec:arch}. In \S\ref{sec:behavior}, we describe how to generate behavior with MSFA.
Finally, we describe the learning algorithm for MSFA in section \S\ref{sec:learning}.

\subsection{Modular Successor Feature Learning}\label{sec:msfl}
Following the Modular RL framework~\citep{russell2003q}, we assume that reward has an additive structure $R(s_t, a_t, s_{t+1}) = \sum_k R^k(s_t, a_t, s_{t+1}) = \sum_k R^{(k)}_t$, where $R^{(k)}_t$ is the reward of the $k$-th module. 
We enforce that every module decomposes reward into an inner-product between its own task-description $w^{(k)}$ and task-agnostic cumulants $\phi^{(k)} \in \mathbb{R}$: $R^{(k)}_t = {\phi^{(k)}_t}\cdot w^{(k)} $.
Here, we simply break up the task vector into $n$ pieces so individual modules are responsible for subsets of the task vector.
This allows us to decompose the action-value function as
\begin{align}
  Q^{\pi}(s_t, a_t, w) = \sum^n_{k=1} Q^{\pi, k}(s_t, a_t, w^{(k)}) = \sum^n_{k=1} \psi^{\pi, k}(s_t, a_t) \cdot w^{(k)}
\end{align}
where we now have \textit{modular SFs} $\{\psi^{\pi,k}(s,a)\}^n_{k=1}$ (see the Appendix for a derivation). Rather than hand-designing modules or cumulants, we aim to discover them from the environment reward signal.

\subsection{Architecture}\label{sec:arch}
We learn a set of modules with states $\mathbb{S}_t = \{s^{(k)}_{t}\}^n_{k=1}$. They update at each time-step $t$ with the observation $x_t$, the previous module-state ${s}^{(k)}_{t-1}$, and information from other modules $A_{\theta}(s^{(k)}_{t-1}, \mathbb{S}_{t-1})$. Following prior work~\citep{santoro2018relational,goyal2019recurrent,carvalho2021feature}, we have $A_{\theta}$ combine transformer-style attention~\citep{vaswani2017attention} with a gating mechanism~\citep{parisotto2020stabilizing} to enforce that inter-module interactions are sparse. Since $A_{\theta}$ is not the main contribution of this paper, we describe these computations in more detail with our notation in the Appendix. We summarize the high-level update below.
\begin{align}\label{eq:update}
  s^{(k)}_t &= s_{\theta_k}(x_t, {s}^{(k)}_{t-1}, A_{\theta}(s^{(k)}_{t-1}, \mathbb{S}_{t-1}))
\end{align}

\textbf{We learn modular cumulants and SFs} by having sets of cumulants and SFs depend on individual module-states. Module cumulants depend on the module-state from the current and next time-step. Module SFs depend on the current module-state and on their subset of the task description. We summarize this below:
\begin{equation}\label{eq:disentangled}
  \tilde{\phi}^{(k)}_t = \phi_{\theta}(s^{(k)}_t, a_t, s^{(k)}_{t+1}) 
  \quad \quad \tilde{\psi}^{w,k}_t = \psi_{\theta}(s^{(k)}_t, a_t, w^{(k)})
\end{equation}
We highlight that cumulants share parameters but differ in their input. This suggests that the key is not having cumulants and SFs with separate parameters but that they are functions of sparse subsets of state (rather than all state information). We show evidence for this hypothesis in~\Figref{fig:borsa_ablation}.

We concatenate module-specific cumulants and SFs to form the final outputs: $\tilde{\phi}_t = \left[ \tilde{\phi}^{(1)}_t, \ldots, \tilde{\phi}^{(n)}_t \right]$ and $\tilde{\psi}^{w}_t = \psi_{\theta}(s_t, a_t, w) = \left[ \tilde{\psi}^{{w},1}_t, \ldots, \tilde{\psi}^{{w},n}_t \right]$. 
Note that cumulants, are only used during learning, update with the module-state from the next time-step. 

\subsection{Behavior}\label{sec:behavior}

During \textbf{training}, actions are selected in proportion to Q-values computed using task SFs as $\pi(s_t, w) \propto {\tilde{Q}(s_t, a, w)} = {{\psi}_{\theta}(s_t, a, w)}^{\top} w$. In practice we use an epsilon-greedy policy, though one can use other choices.
During \textbf{testing}, we compute policies with GPI as $\pi(s_t, w_{\tt test}) \in \argmax_{a} \max_{z \in \mathbb{M}_{\tt train}} \{ { {\psi}_{\theta}(s_t, a, z)}^{\top} w_{\tt test} \}$, where $\mathbb{M}_{\tt train}$ are train task vectors. 

\subsection{Learning Algorithm}\label{sec:learning}

MSFA relies on three losses. The first loss, $\mathcal{L}_Q$, is a standard Q-learning loss, which MSFA uses to learn optimal policies for the training tasks. The main difference here is that MSFA uses a particular parameterization of the Q-function $Q^{\pi_w, w}(s,a) = \psi^{\pi_w}(s,a)^{\top}w$. The second loss, $\mathcal{L}_\psi$, is an SF learning loss, which we use as a regularizer to enforce that the Q-values follow the structure in the reward function $r^w_t = \phi_t^{\top} w$. For this, we again apply standard Q-learning but using SFs as value functions and cumulants as pseudo-rewards. The final loss, $\mathcal{L}_{\phi}$ is a loss for learning cumulants that grounds them in the environment reward signal. The losses are summarised as follows
\begin{align} \label{eq:losses}
  \mathcal{L}_Q &= ||r_t + \gamma\psi_{\theta}(s_{t+1}, a', w)^{\top}w - \psi_{\theta}(s_{t}, a_t, w)^{\top}w ||^2 \\
  \mathcal{L}_{\psi} &= || {\tilde{\phi_t}} + \gamma\psi_{\theta}(s_{t+1}, a', w) - \psi_{\theta}(s_{t}, a_t, w) ||^2 \\
  \mathcal{L}_{\phi} &= || r^w_t - \tilde{\phi_t}^{\top} w ||^2
\end{align}
where $a' = \argmax_a \psi_{\theta}(s_{t+1}, a, w)^{\top}w$. Selecting the next action via the combination of all modules ensures they individually convergence to optimal values~\citep{russell2003q}.
The final loss is $\mathcal{L} = \alpha_{Q} \mathcal{L}_Q + \alpha_{\psi} \mathcal{L}_{\psi} + \alpha_{\phi} \mathcal{L}_{\phi}$.

\section{Experiments}\label{sec:exp}
\begin{figure}[!htb]
  \centering
  \includegraphics[width=.7\textwidth]{\msfapath/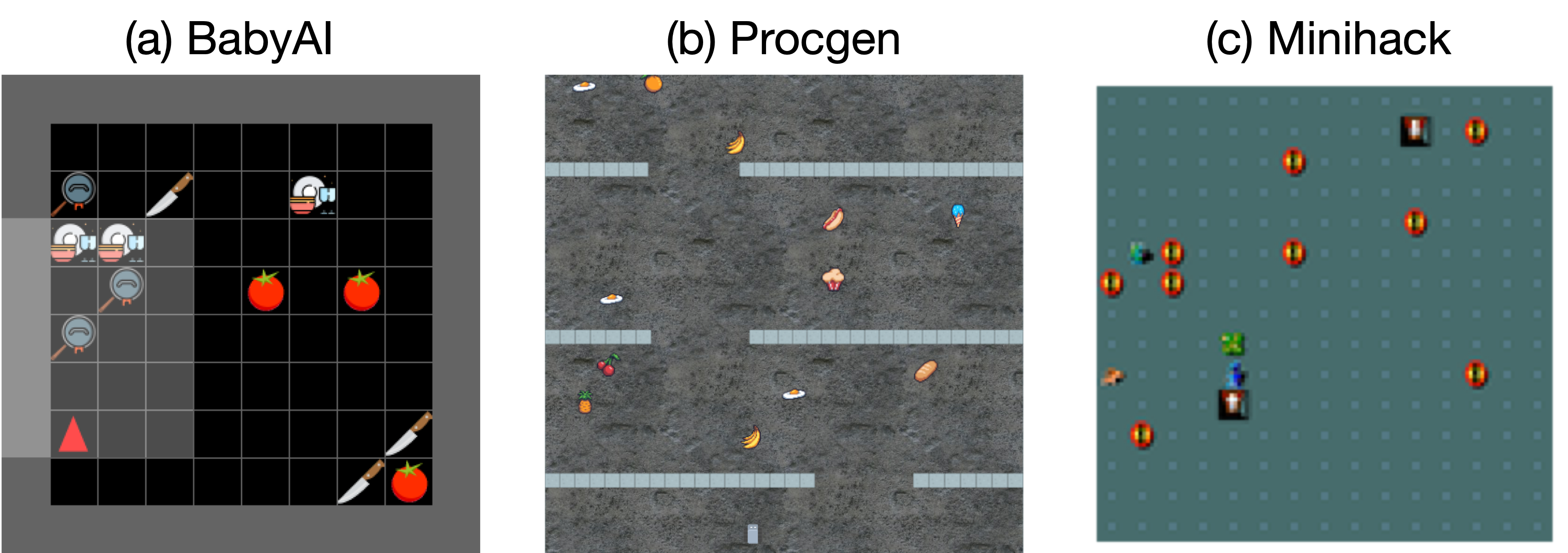}
  \caption{\textbf{We study an agent's ability to combine task knowledge in three environments.} (a) In BabyAI, an agent learns to pick up one object type at a time during training. During testing, the agent must pickup combinations of object types while avoiding other object types. This is the setting used by USFA which assumed \textbf{hand-designed} cumulants. (b) In Procgen, we study extending this form of generalization to a visually diverse, procedurally generated environment. (c) In Minihack, we go beyond combining object navigation skills. Here, an agents needs to combine (1) avoiding teleportation traps, (2) avoiding monsters, and (3) partial visibility around the agent.
    }\label{fig:envs}
\end{figure}

We study generalization when training behaviors must be combined concurrently in time in the presence of novel object appearances and layouts. The need to combine training behaviors tests how well MSFA can leverage SF\&GPI. Generalization to novel object appearances and layouts tests how well MSFA's modular construction supports generalization to novel environment configurations.

\textbf{Baselines.}
(1) \textbf{Universal Value Function Approximator (UVFA)}~\citep{schaul2015universal}, which takes the task as input: $Q_{\theta}(s, w)$. This comparison shows shows the transfer benefits of SF\&GPI.
(2) \textbf{UVFA with Feature-Attending Recurrent Modules (UVFA+FARM)} instead takes state-factors as input $Q_{\theta}(s^{(1)}, \ldots, s^{(n)}, w)$. Each state-factor $s^{(k)}$ is the output of a FARM module.
(3) \textbf{Modular Value Function Approximator(MVFA)} is an adaptation of~\citep{haarnoja2018composable} where modules learn individual Q-values $Q^{(i)}_{\theta}(s^{(i)}, w^{(i)})$.
Comparing to UVFA+FARM and MVFA enables us to study the benefits of leveraging modules for learning value functions in the form of SFs.
(4) \textbf{Universal Successor Function Approximator (USFA)}~\citep{borsa2018universal} leverages a single monolithic function for successor features with \textbf{hand-designed cumulants}.
USFA is an upper-bound baseline that allows us to test the quality of cumulants and successor features that MSFA learns.
We also test a variant of USFA with learned cumulants, \textbf{USFA-Learned-$\phi$}, which   shows how the architecture degrades without oracle cumulants. 

\textbf{Implementation.}
We implement the state modules of MSFA with FARM modules~\citep{carvalho2020reinforcement}.
For UVFA and USFA, we learn a state representation with an LSTM~\citep{hochreiter1997long}.
We implement all $\phi$, $\psi$, and $Q$ functions with Mutli-layer Perceptrons.
We train UVFA and UVFA+FARM with n-step Q-learning~\citep{watkins1992q}. 
When learning cumulants, USFA and MSFA have the exact same losses and learning alogirthm. They both learn Q-values and SFs with n-step Q-learning. We use $n=5$.
When not learning cumulants, following~\citep{borsa2018universal}, USFA only learns SFs with n-step Q-learning.
All agents are built with JAX~\citep{jax2018github} using the open-source ACME codebase~\citep{hoffman2020acme} for reinforcement learning.

\subsection{Combining object navigation task knowledge}\label{sec:exp-babyai}
\begin{figure}[!htb]
  \centering
  \includegraphics[width=.85\textwidth]{\msfapath/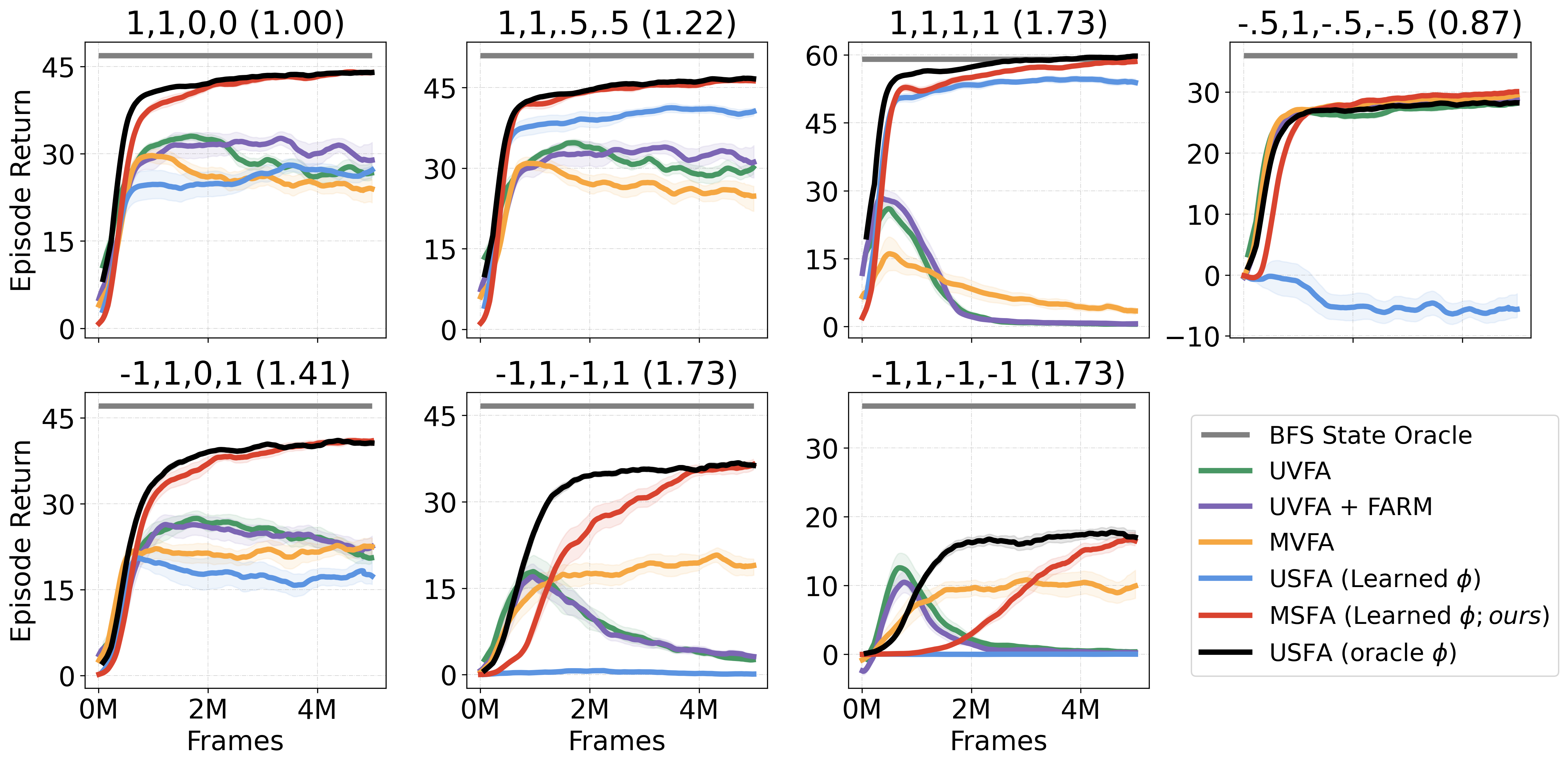}
  \caption{ \textbf{MSFA matches USFA, which has hand-designed cumulants}.
  We show mean and standard error generalization episode return across 10 runs. We put a task's L2 distance to the closest train task in parenthesis.
  USFA best generalizes to novel combinations of picking up and avoiding objects. Once USFA learns cumulants, its performance degrades significantly. UVFA-based methods struggle as more objects should be avoided or tasks are further in distance to train tasks.
  }\label{fig:borsa_main}
\end{figure}

MSFA learns modular functions for computing $\phi$ and estimating $\psi$. We hypothesize that this facilitates learning cumulants that respond to different aspects of the environment (e.g.~to different object categories). This leads to the following research questions.
\textbf{R1}: Can we recover prior generalization results that relied on hand-designed cumulants for different object categories?
\textbf{R2}: How important is it to learn modular functions for $\phi$ and $\psi$?
\textbf{R3}: Without GPI, do learning modular functions for $\phi$ and $\psi$ still aid in generalization?

\textbf{Setup.} We implement a simplified version of the object navigation task of~\citep{borsa2018universal} in the~\citep{babyai_iclr19} BabyAI environment.
The environment consists of $3$ instances of $4$ object categories. 
\textbf{Observations} are partial and egocentric.
\textbf{Actions}: the agent can rotate left or right, move forward, or pickup an object. When it picks up an object, following~\citep{borsa2018universal}, the object is respawned somewhere on the grid. 
\textbf{Task} vectors lie in $w\in \R^4$ with training tasks being the standard unit vectors. For example, $[0,1,0,0]$ specifies the agent must obtain objects of category $2$.
\textbf{Generalization} tasks are linear combinations of training tasks. For example, $[-1,1,-1,1]$ specifies the agent must obtain categories $2$ and $4$ while \textit{avoiding} categories $1$ and $3$.
~\citet{borsa2018universal} showed that USFA could generalize with \textit{hand-designed} cumulants that described whether an object was picked up. 
We describe challenges for this task in detail in the Appendix.

\textbf{R1: MSFA is competitive with USFA, which uses oracle $\phi$.} \Figref{fig:borsa_main} shows USFA with a similar generalization trend to~\citep{borsa2018universal}. Tasks get more challenging as they are further from train tasks or involve avoiding more objects. For simply going to combinations of objects, USFA-Learned-$\phi$ does slightly worse than MSFA. However, with more objects to avoid, all methods except MSFA (including USFA-Learned-$\phi$) degrades significantly. For comparison, we show performance by an oracle bread-first-search policy with access to ground-truth state (\textbf{BFS State Oracle}). All methods have room for improvement when objects must be avoided. In the appendix, we present heat-maps for how often object categories were picked up during different tasks. We find that MSFA most matches USFA, while USFA-Learned-$\phi$ commonly picks up all objects regardless of task.

\begin{figure}[!htb]
  \centering
  \vspace{-3pt}
  \includegraphics[width=.95\textwidth]{\msfapath/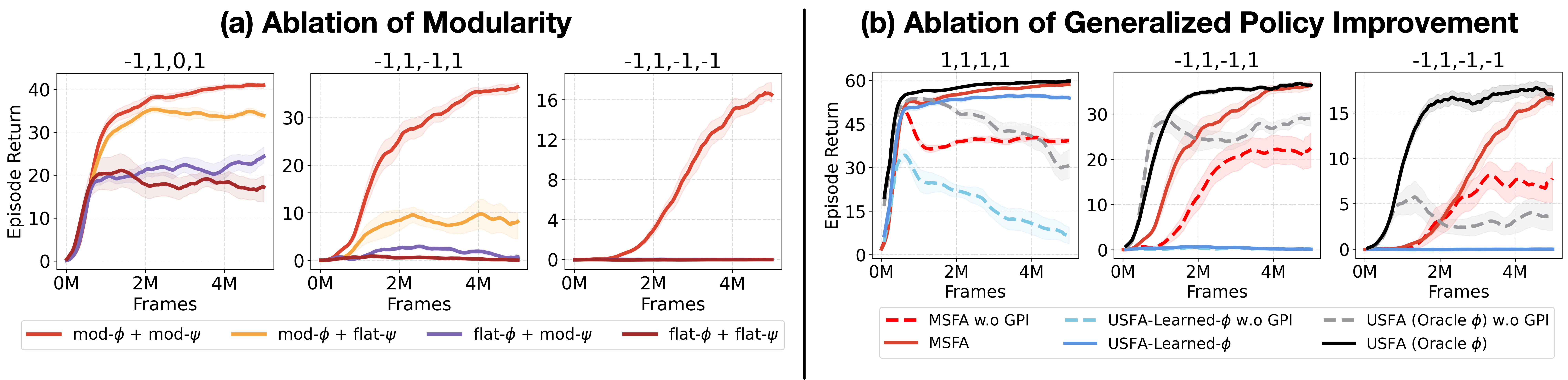}
  \caption{ \textbf{Learning modular $\phi$ and $\psi$ is key to generalization and improves generalization even without GPI.}
  We show mean and standard error generalization episode return across 10 runs.
  (a) We ablate having modular functions for $\phi_{\theta}$ and $\psi_{\theta}$. Generalization results degrade significantly. (b) We ablate leveraging GPI for generalization from all SF-based methods. MSFA without GPI can outperform both USFA-Learned-$\phi$ with GPI and USFA without GPI. This shows the utility of modularity for generalization.}\label{fig:borsa_ablation}
  \vspace{-10pt}
\end{figure}

\textbf{R2: Learning modular $\phi$ and $\psi$ functions is critical for generalization.}
Learning an entangled function corresponds to learning a monolithic function for $\psi$ or $\phi$ where we concatenate module-states, e.g. $\tilde{\psi}^{w}_t=\psi_{\theta}(s^{(1)}_t, \ldots, s^{(n)}_t, a, w)$.
Modular functions correspond to~\eqref{eq:disentangled}.
\Figref{fig:borsa_ablation} (a) shows that without modular functions for $\phi$ \textbf{and} $\psi$, performance severely degrades.
This also highlights that a naive combination of USFA+FARM---with entangled functions for $\phi$ and $\psi$---does not recover our generalization performance.

\textbf{R3: Modularity alone improves generalization.} For all SF-based methods, we remove GPI and select actions with a greedy policy: $\pi(s) = \argmax_a \psi_{\theta}{(s_t, a, w)}^{\top}w$. 
\Figref{fig:borsa_ablation} (b) shows that GPI is critical for generalization with USFA as expected. USFA-Learned-$\phi$ benefits less from GPI (presumably because of challenges in learning $\phi$). Interestingly, MSFA can generalize relatively well without GPI, sometimes doing better than USFA without GPI.

\subsection{Combining object navigation task knowledge with novel appearances and environment configurations}\label{sec:exp-procgen}
Beyond generalizing to combinations of tasks, agents will need to generalize to different layouts and appearances of objects. 
\textbf{R4}: Can MSFA enable combining task knowledge in a visually diverse, procedurally generated environment?

\textbf{Setup.}  We leverage the ``Fruitbot'' environment within ProcGen~\citep{cobbe2020leveraging}.
Here, an agent controls a paddle that tries to obtain certain categories of objects while avoiding others. When the agent hits a wall or fence, it dies and the episode terminates. If the agent collects a non-task object, nothing happens.
\textbf{Observations} are partial.
\textbf{Actions}: At each time-step the agent moves one step forward and can move left or right or shoot pellets to open fences.
\textbf{Training and generalization tasks} follow the same setup as \S\ref{sec:exp-babyai}.
\begin{figure}[!htb]
  \centering
  \includegraphics[width=.9\textwidth]{\msfapath/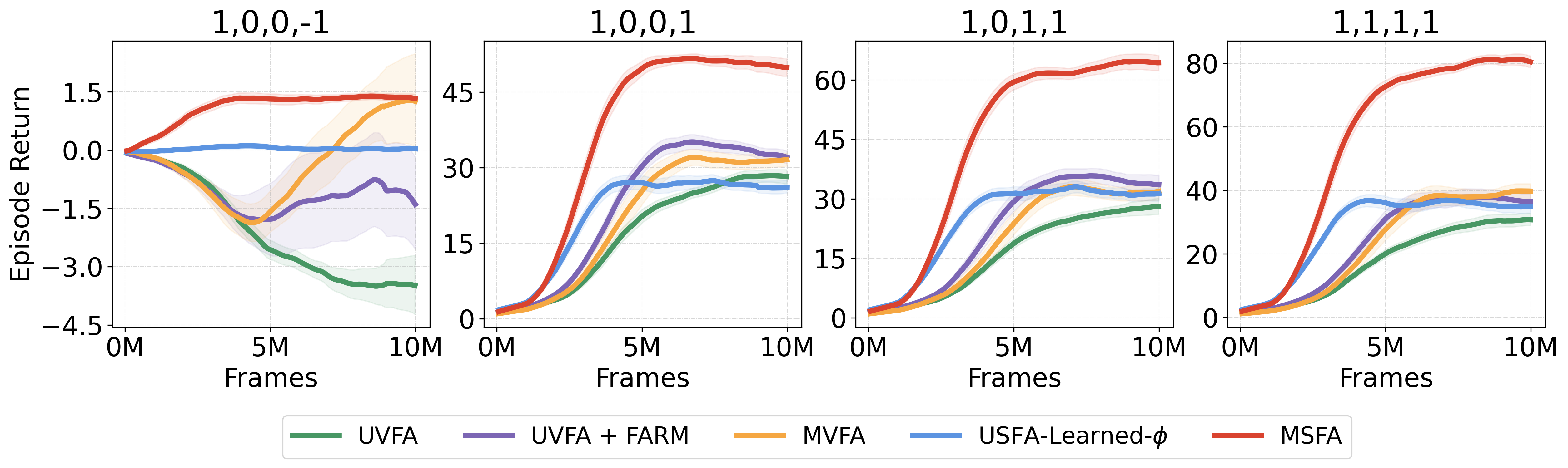}
  \caption{\textbf{MSFA is able to combine task knowledge in a visually diverse, procedurally generated ProcGen environment.} We find that no method is able to do well when there are objects to avoid ($w=[1, 0, 0, -1]$) in this setting (see text for more). However, as more objects need to be collected MSFA best generalizes (10 runs). 
    }\label{fig:fruitbot_taskgen}
\end{figure}

\textbf{R4: MSFA enables combining task knowledge of object navigation tasks in a visually diverse, procedurally generated environment.} \Figref{fig:fruitbot_taskgen} shows that when an agent has to generalize to collecting more objects, modular architectures generalize best, with MSFA doing best. When objects have to be avoided, we see that no architecture does well, though MSFA tends to do better. 
We observe that avoiding objects leads agents to hit walls. 
Since the agent always moves forward at each time-step in Fruibot, this makes avoiding objects a particularly challenging type of generalization.

\section{Discussion and Conclusion}

We have presented ``Modular Successor Feature Approximators'', a modular neural network for learning cumulants and SFs produced by their own modules. 
We first showed that MSFA is competitive with prior object navigation generalization results that relied on hand-designed cumulants (\S\ref{sec:exp-babyai}).
Afterwards, we showed that MSFA improves an agent's ability to combine task knowledge in a visually diverse, procedurally generated environment (\S\ref{sec:exp-procgen}).
We also show that MSFA can combine solutions of heterogeneous tasks (\S\ref{sec:exp-minihack}).
Our ablations show that learning modular cumulants and SFs is critical for generalization with GPI. 

We compared MSFA to
(1) USFA, a monolithic architecture for learning cumulants and SFs;
(2) FARM, an architecture which learns multiple state-modules but combines them with a monolithic Q-value function. Our results show that when learning cumulants, MSFA improves generalization with SF\&GPI compared to USFA. Additionally, without GPI, MSFA as an architecture improves generalization as compared to both FARM and USFA.

\textbf{Limitations}. While we demonstrated reward-driven discovery of cumulants for transfer with SF\&GPI, we focused on relatively simple task encodings. Future work can extend this to more expressive encodings such as language embeddings.
Another limitation is that we did not explore more sophisticated state-feature discovery methods such as meta-gradients~\citep{veeriah2019discovery}.
Nonetheless, we think that MSFA provides an important insight for future work: that modularity is a simple but powerful inductive bias for discovering state-features that enable zero-shot transfer with SF\&GPI.

\textbf{Future directions}. SFs are useful for exploration~\citep{janz2019successor, machado2020count}; for discovering and combining options~\citep{barreto2019option, hansen2019fast}; for transferring policies across environments~\citep{zhang2017deep}; for improving importance sampling~\citep{fujimoto2021deep}; and for learning policies from other agents~\citep{filos2021psiphi}. We hope that future work can leverage MSFA for improved state-feature discovery and SF-learning in all of these settings.

\section*{Acknowledgments}
This work was supported in part by a University of Michigan Rackham Merit Fellowship and grant from LG AI Research. The authors would also like to thank Andrew Lampinen, Cameron Allen, and the anonymous reviewers for their helpful comments.

\bibliography{bib}
\bibliographystyle{iclr2023_conference}

\clearpage
\appendix

\section{Learning and Generalization Challenges}\label{appendix:challenges}
Our experimental setup for \S\ref{sec:exp-babyai} follows the same setup as \citet{barreto2018transfer}. For $n$ tasks with reward functions $\{r_i\}^n_{i=1}$, they showed that discovered features which predict rewards, i.e. $[\tilde{\phi}_1, \ldots, \tilde{\phi}_n]= [\tilde{r}_1, \ldots, \tilde{r}_n]$, were effective features for transfer with SF\&GPI. In their experimental setup, training tasks were one-hot vectors, so regressing $||r-\tilde{\phi}^{\top}w||$ effectively led $\phi_i$ to be a feature that predicted reward $r_i$. We use this same setup, where now \textit{individual modules} are used for both $\tilde{\phi}^{(i)}$ and $\tilde{\psi}^{(i)}$.

\begin{wrapfigure}[26]{R}{0.45\textwidth}
  \vspace{-10pt}
  \centering
  \includegraphics[width=.45\textwidth]{\msfapath/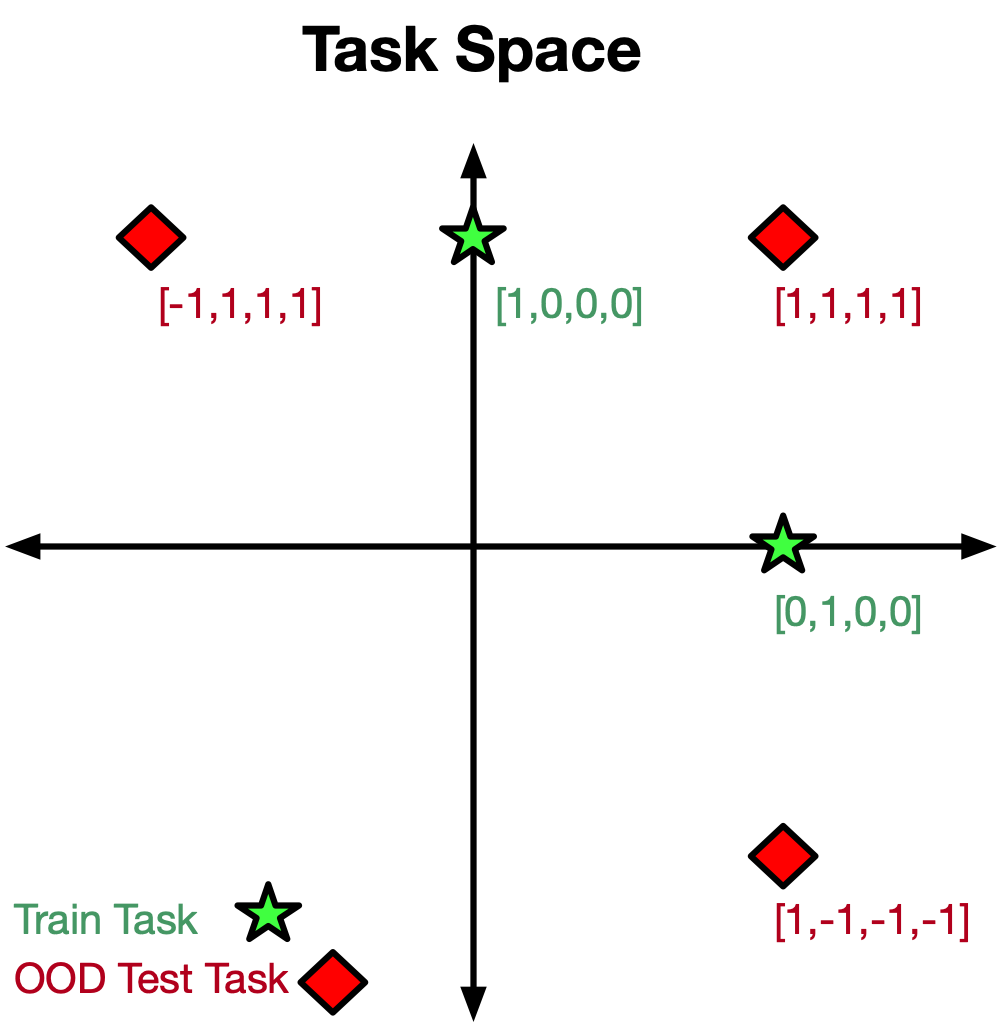}
  \caption{We present a diagram providing intuition for the type of generalization challenge required for our tasks. Green stars represent our training tasks. Red diamonds represent test tasks. They are both outside of training distribution for task encodings and can be far in task-space from training task encodings.
  }\label{fig:task-ood}
\end{wrapfigure}
While this is a simple method for learning features, it some difficulties. First, there is an imbalance of rewarding vs. non-rewarding transitions, which leads to challenges in learning $\phi$. In our setting, an \textit{optimal policy} only has $7\%$ of its transitions be rewarding transitions. Over the course of training, the percentage is much lower. Second, \citet{barreto2018transfer} were mainly able to show \textit{continual learning} results with additional learning using their discovered features and had \textbf{limited success with zero-shot transfer}.

Zero-shot transfer is challenges because training tasks are ``out-of-distribution'' while additionally requiring behaviors that the agent was never trained on. By out-of-distribution, we mean that the agent is only trained on basis vectors (see \Figref{fig:task-ood}). However, at test time, the agent must perform tasks that are are far in ``task-space'' from the training distribution. For example, $w_{\tt test} = [-1, 1, -1, -1]$ has an L2 distance of $1.73$ from closest training task.
This requires an agent that can generalize to test reward functions that are quite different from training reward functions. While both USFA-Learned-$\phi$ and MSFA get about the same training error for predicting rewards, we see that MSFA gets a dramatically better reward prediction error for test reward functions $\tilde{r}_{\tt test}= \tilde{\phi}^{\top}w_{\tt test}$.

\textbf{Reward prediction error on generalization tasks}. To investigate this, we collect 40 episodes of each generalization task and compute the reward prediction error of MSFA and USFA-Learned-$\phi$ for each task. We present the results in Table~\ref{table:prediction-error}.

\begin{table}[!htp]\centering
\begin{tabular}{l|l|l}
    \toprule
             \textbf{Task} &                \textbf{MSFA Error} &                \textbf{USFA-Learned-$\phi$ Error} \\
             \midrule
          1,1,0,0 &     $0.0003 \pm     0.0090$ &     $0.1362 \pm     0.3410$ \\
        1,1,.5,.5 &     $0.0069 \pm     0.0405$ &     $0.3922 \pm     0.7051$ \\
          1,1,1,1 &     $0.0118 \pm     0.0856$ &     $2.5530 \pm     4.0324$ \\
    -.5,1,-.5,-.5 &     $0.0095 \pm     0.0478$ &     $0.0556 \pm     0.1041$ \\
         -1,1,0,1 &     $0.0014 \pm     0.0264$ &     $0.0652 \pm     0.2470$ \\
    \bottomrule
  \end{tabular}
\caption{\textbf{MSFA has a smaller reward prediction error for test tasks}. We present the reward prediction error for MSFA and USFA-Learned-$\phi$. We present results using 40 episodes for each test task.}\label{table:prediction-error}
\end{table}

\section{Combining knowledge of heterogeneous tasks}\label{sec:exp-minihack}
Aside from navigation tasks, researchers and practitioners will be interested in generally combining solutions to different tasks. \textbf{R5}: Can MSFA enable combining solutions to heterogeneous tasks?

\begin{figure}
  \centering
  \includegraphics[width=.45\textwidth]{\msfapath/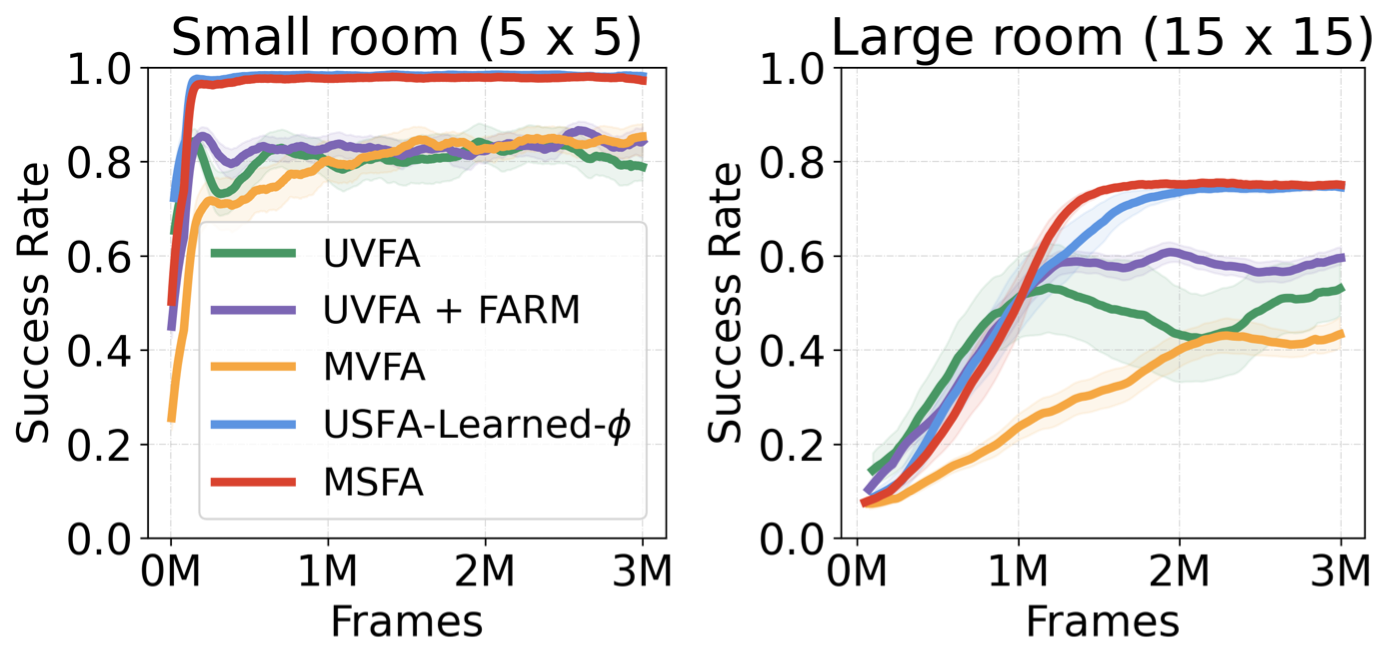}
  \caption{We study whether MSFA can combine three separate settings (1) limited visibility (2) the presence of monsters (3) the presence of teleportation traps (see text for more). {All successor feature based methods best enable combining training tasks for generalization in this setting.}
  (10 run)}\label{fig:minihack}
\end{figure}

\textbf{Setup.}  We leverage the ``Minihack'' environment~\citep{samvelyan2021minihack}. Here, an agent is spawned at the top of a map with obstacles. It must navigate to the a ladder at the bottom of the map.
The agent experiences partial \textbf{observations} of the environment.
\textbf{Actions}: At each time-step the agent can select from $8$ compass directions.
\textbf{Training tasks}: The agent experiences three training tasks (1) avoiding monsters which kill the agent, (2) avoiding traps which teleport the agent away from the goal, and (3) experiencing a small local view of the environment with everything else black. \textbf{Generalization} requires that the agent generalize to (a) new configurations of objects and (b) to the \textit{combination} of all training conditions. 

\textbf{R5: MSFA enables combining solutions to heterogeneous tasks.} \Figref{fig:minihack} shows that SF-based methods best combine solutions to different tasks. In a small room, both SF-based methods get 100\% success rate. In a larger room, no method generalizes perfectly: SF-based methods do best at 75\%, followed by UVFA-FARM and UVFA with 60\%.

\section{Derivation of Modular Successor Features}\label{appendix:deriv}
\begin{align}\label{eq:sf_deriv}
  Q^{\pi}(s, a, w) 
  &= \mathbb{E}_{\pi}\left[ \sum^\infty_{j=0} \gamma^j R_t |S_t=s, A_t=a\right] \\
  &= \mathbb{E}_{\pi}\left[ \sum^\infty_{j=0} \gamma^j\left( \sum^n_{k=1} {\phi_{t+j}^{(k)}}^{\top}w^{(k)}\right) |S_t=s, A_t=a\right] \\
  &= \sum^n_{k=1} \left(\mathbb{E}_{\pi}\left[ \sum^\infty_{j=0} \gamma^j  {\phi_{t+j}^{(k)}}^{\top}w^{(k)} |S_t=s, A_t=a\right] \right) \\
  &= \sum^n_{k=1} \left(\mathbb{E}_{\pi}\left[ \sum^\infty_{j=0} \gamma^j  {\phi_{t+j}^{(k)}} | S_t=s, A_t=a\right]^{\top}w^{(k)} \right) \\
  &= \sum^n_{k=1} \psi^{\pi, k}(s, a)^{\top} w^{(k)} \\
  &= \sum^n_{k=1} Q^{\pi,k}(s, a, w^{(k)})
\end{align}

\section{Inter-module attention and gating}\label{appendix:attn}
At each time-step $t$, each module updates with both observation features $z_t = f^{\theta}_z(x_t)$ and with information from the previous module-states $\{s^{(k)}_{t-1} \}$. MSFA shares information between modules with a ``query-key'' system. Each module has a ``query'' representation of its module-state that it uses to select from ``key'' representations of the other module-states. MSFA scores how well a query matches a key by computing the dot-product between the two representations. A module then updates with the module-state corresponding to the key with the highest dot-product. To enable flexible updating, each module uses a gating mechanism to decide the degree to which information should be incorporated into an update.

More technically, the query vector is computed using the previous-module state and action: $q^{(k)}_t = W^{\tt query}[s^{(k)}_{t-1}, a_{t-1}] \in \mathbb{R}^{d_q}$.
Keys and values are computed as $K_t = W^{\tt key}[s_{t-1}^1; \ldots; s_{t-1}^n; 0] \in \mathbb{R}^{n+1 \times d_q}$ and $V_t = W^{\tt value}[s_{t-1}^1; \ldots; s_{t-1}^n; 0] \in \mathbb{R}^{n+1 \times d_q}$, respectively. Note that we add a zero-vector key and value to allow a query to select ``no information''. Each module selects ``values'' $v^{(k)}_t$ to update using dot-product attention~\citep{vaswani2017attention}:
\begin{align}
  v^{(k)}_{t} = \softmax\left(\frac{q_t^{(k)} K^{\top}_t}{d_q}\right)V_t
\end{align}
Gating mechanism has been shown important for leveraging transformer-style attention when updating state in RL agents~\citep{parisotto2020stabilizing}. Thus, each module uses a sigtanh gate when updating:
\begin{equation}
  u^{(k)}_{t} = q^{(k)}_{t} + \tanh(W^{g_1} v^{(k)}_{t}) \odot \sigma(W^{g_2} v^{(k)}_{t} - b_g).
\end{equation}
MSFA then updates its module-stateas follows:
\begin{equation}
  s^{(k)}_t = s_{\theta_k}(u^{(k)}_{t}, z_t) 
\end{equation}

\section{Learning modular functions}\label{appendix:modular}
Our goal is to learn functions for producing cumulants $\phi$ and SFs $\psi$ that have consider only state information from their own modules. To be more precise, we learn modules which collectively learn a set of $n$ module-state representations: $\{s^{i}\}^{n=1}_i$. As an example, consider learning a linear function/transformation $A$ for producing cumulants. A typical ``monolithic function'' would be one like the following:
$$
\begin{bmatrix}
         \phi^1 \\
         \vdots \\ 
         \phi^n 
 \end{bmatrix}
 = 
 \begin{bmatrix}
         A_{11} & \cdots & A_{1m}\\
         \vdots & A_{22} & \vdots\\ 
         A_{n1} & \cdots & A_{nm} 
 \end{bmatrix}
 \begin{bmatrix}
         s^1 \\
         \vdots \\ 
         s^n 
 \end{bmatrix}.
$$
By "entangled," we mean that a cumulant $\phi^i$ is (potentially) a function of all module-state information  $\phi^i=\sum_j A_{ij} s^j$. By a "modular function", we simply meant that we were learning a function like the following
$$
\begin{bmatrix}
         \phi^1 \\
         \vdots \\ 
         \phi^n 
 \end{bmatrix}
 = 
 \begin{bmatrix}
         A_{1} & 0 & 0\\
         0 & A_{2} & 0\\ 
         0 & 0 & A_{n} 
 \end{bmatrix}
 \begin{bmatrix}
         s^1 \\
         \vdots \\ 
         s^n 
 \end{bmatrix}.
$$
In our setting, we had $A_1=\ldots=A_n$ be a shared MLP and each $s^i$ was produced by a module with its own parameters. Our experiments demonstrate leveraging modules to learn $\phi^{(i)}$ and $\psi^{(i)}$ is an effective way to learn $\{\phi^i\}_i$ that promote zero-shot composition of task knowledge with the SF\&GPI framework.

\section{Additional analysis}\label{appendix:analysis}
\begin{figure}[!htb]
  \centering
  \includegraphics[width=.6\textwidth]{\msfapath/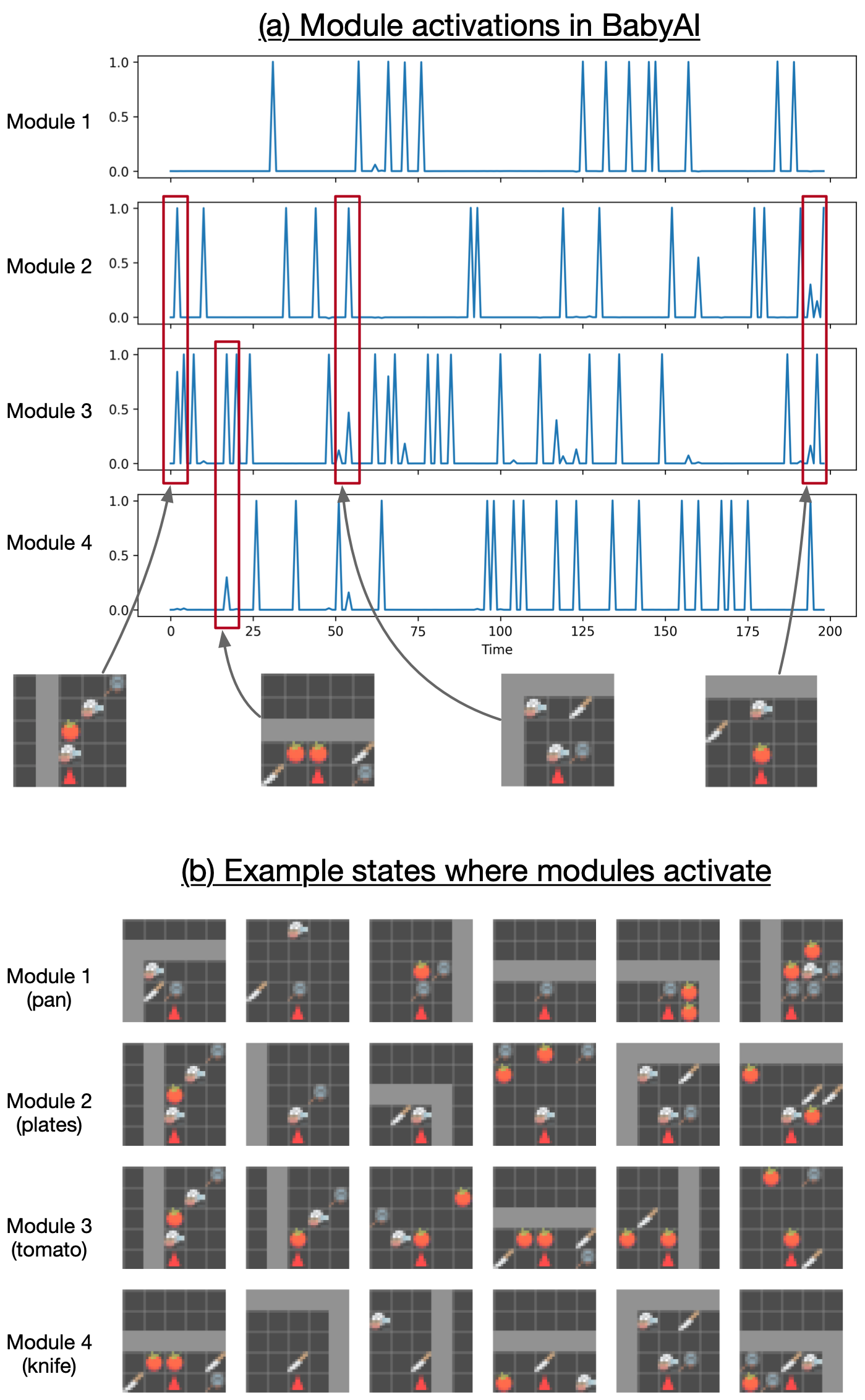}
  \caption{We present a visualization of module activity for task generalization task $w=[1,1,1,1]$ from \S\ref{sec:exp-babyai}. On the top, (a) we show the activity of each module across time. On the bottom, (b) we show example transitions where modules activate to visualize what they respond to. We see that individual modules are able to effectively respond to different object categories, despite a strong data imbalance involved in learning to represent each object category (see \S\label{appendix:challenges}). We find that modules don't perfectly specialize, and multiple modules will sometimes activate at the same state.
  }\label{fig:visualization}
\end{figure}

\begin{figure}[!htb]
  \centering
  \includegraphics[width=\textwidth]{\msfapath/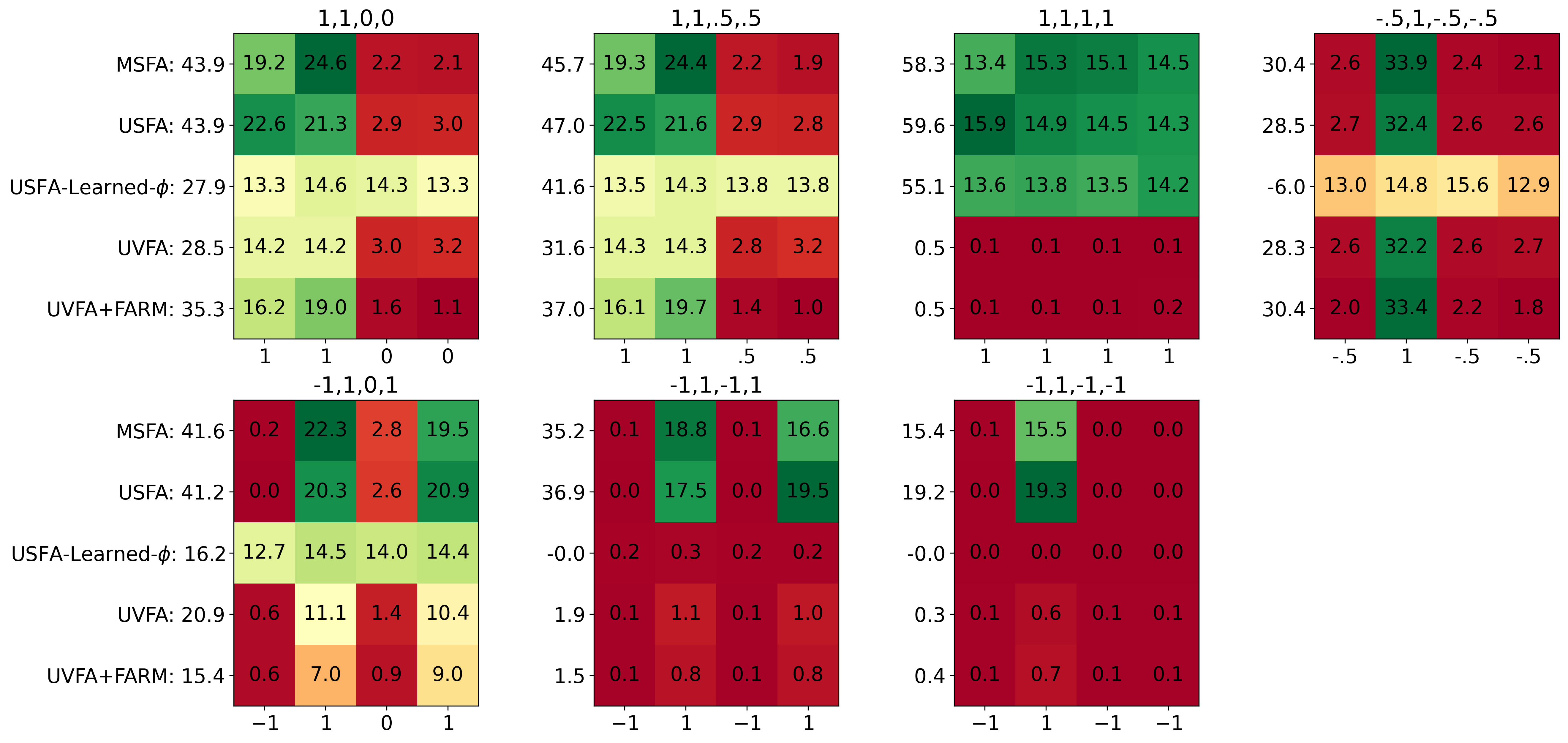}
  \caption{\textbf{MSFA best matches the behavior of USFA across generalization tasks}. We present additional analysis for the experiments in \S\ref{sec:exp-babyai}. Each heatmap displays how often object categories were picked up by each method for a given generalization task. Rows correspond to different methods. We show the final episode return for each method along the y-axis. Columns describe the transition-reward when an object category is picked up. USFA has access to hand-designed cumulants that described whether an object category was picked up. Despite learning cumulants, MSFA best matches USFA's object collection dynamics. USFA-Learned-$\phi$ equally collects all objects regardless of task, except for when $r\leq-1$ when an object is picked up. In this setting, no method except USFA or MSFA collect objects. As a reminder, negative rewards are never experienced during training and these tasks are the furthest away from training tasks in task-space. UVFA-based methods tend to collect rewarding objects but not as effectively as MSFA or USFA.
  }
  \label{fig:behavior}
\end{figure}

\begin{figure}[!htb]
  \centering
  \includegraphics[width=.7\textwidth]{\msfapath/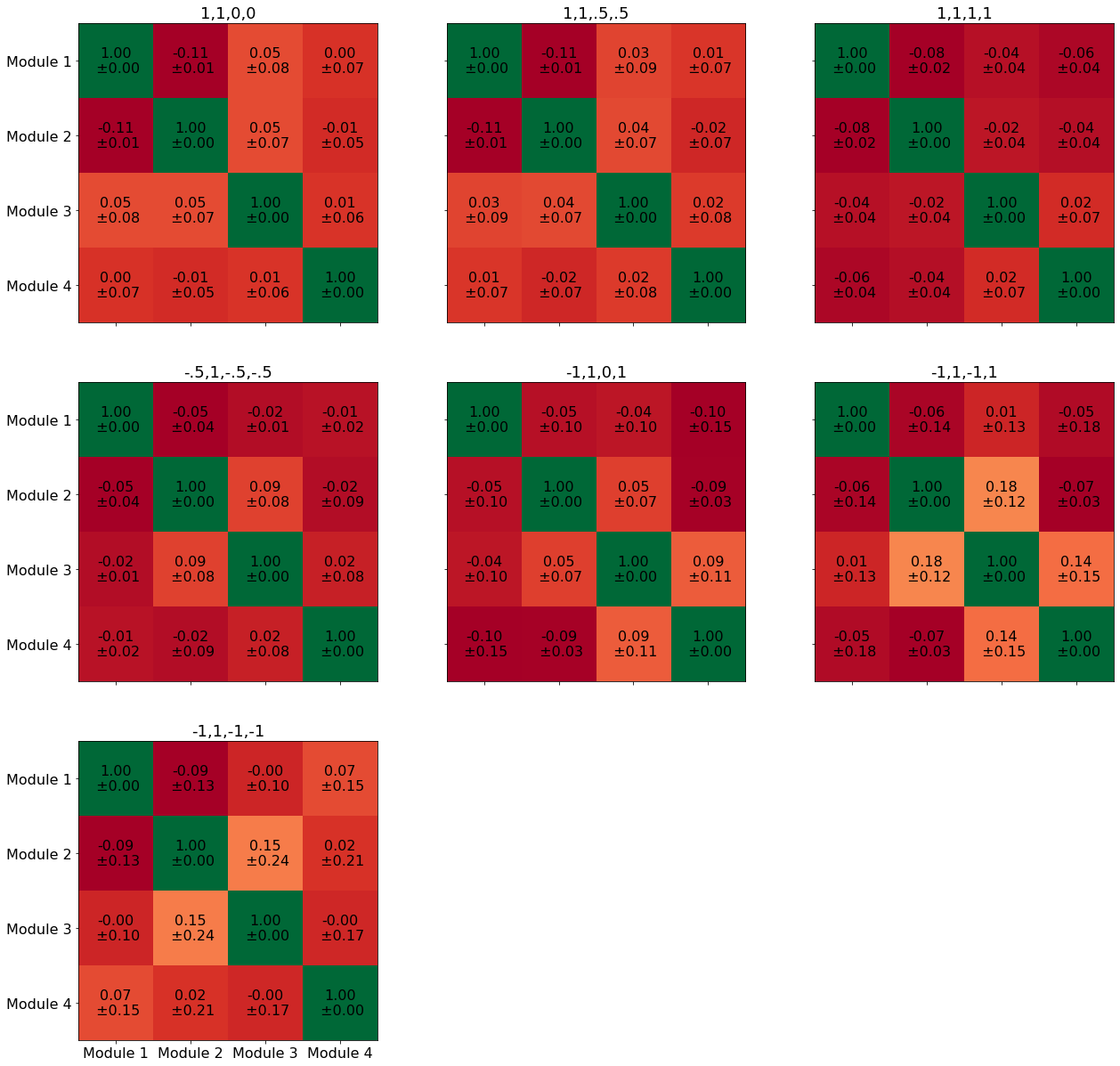}
  \caption{We present the cross-correlation between pairs of modules. This confirms that modules tend to activate at different time-points on a statistical level.}
  \label{fig:correlation}
\end{figure}

\clearpage
\section{Full results}\label{appendix:results}
\begin{figure}[!htb]
  \centering
  \includegraphics[width=\textwidth]{\msfapath/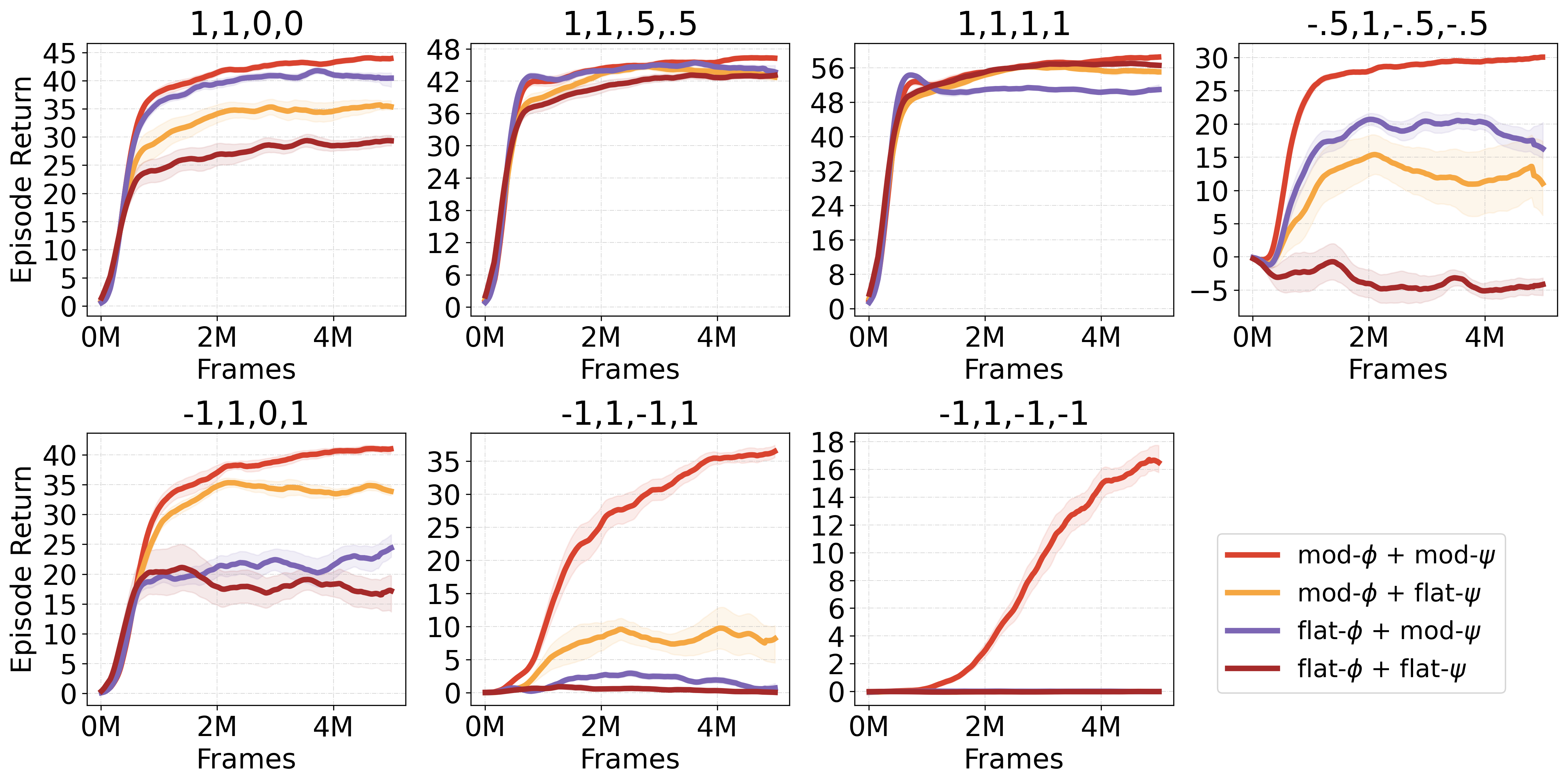}
  \caption{ All results for ablating disentanglement of $\phi$ and $\psi$. We consistently find that disentangled functions for $\phi$ and $\psi$ have the best generalization performance.
  }
\end{figure}

\begin{figure}[!htb]
  \centering
  \includegraphics[width=\textwidth]{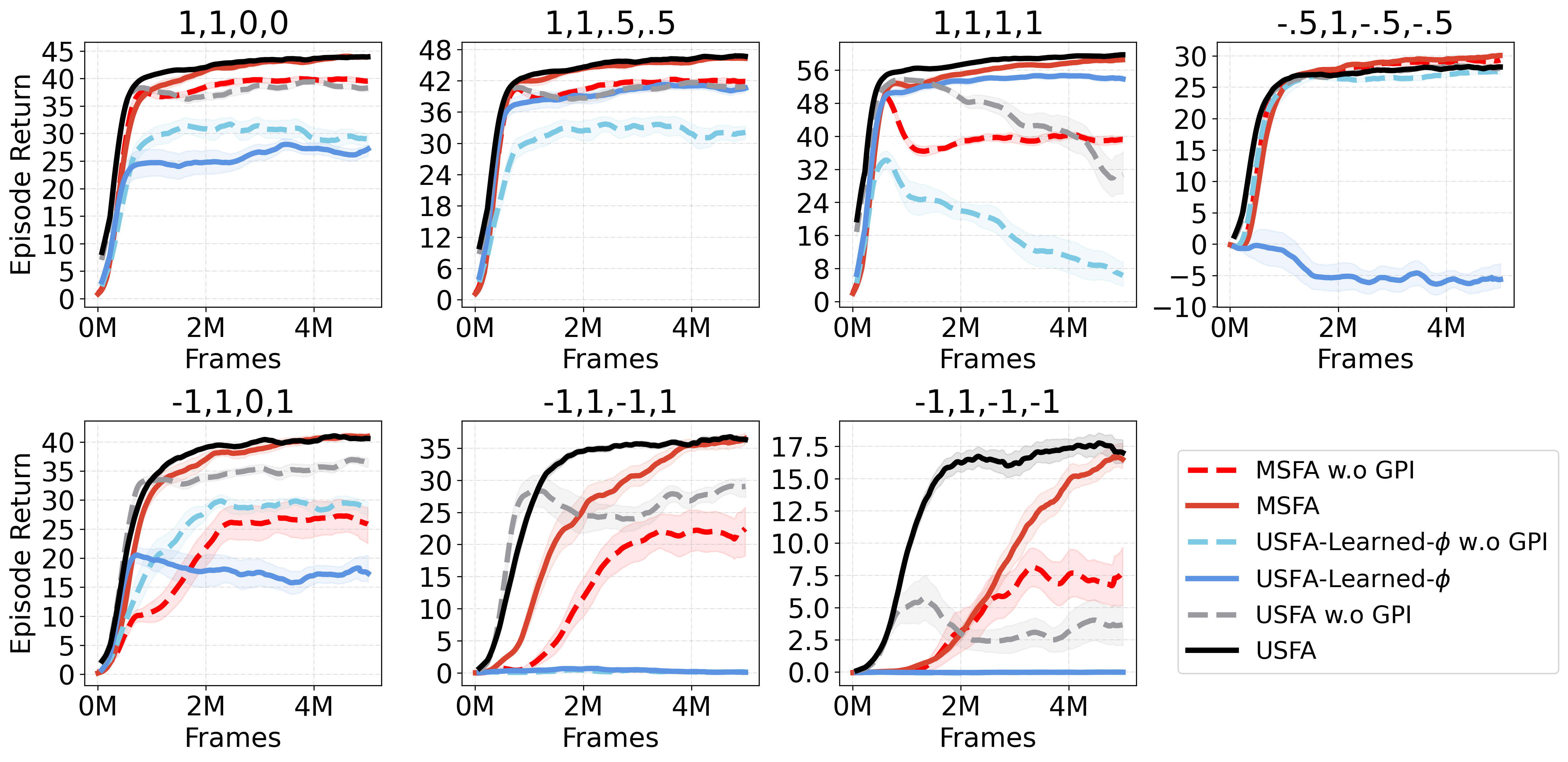}
  \caption{All results for ablating GPI. First, we compare MSFA to USFA (which has hand-designed cumulants). We can see that the two perform comparably across the board. Sometimes (e.g. $[1,1,1,1]$ and $[-1,1,-1,-1]$), MSFA outperforms USFA. Next we compare MSFA to USFA-Learned-$\phi$. We see that MSFA performs as well as USFA-Learned-$\phi$ or better in all settings except $[-1,1,0,1]$. Interestingly, MSFA without GPI outperforms USFA-Learned-$\phi$ with GPI in some generalization settings (e.g. $[1,1,0,0]$ and $[-1,1,-1,1]$).
  }
\end{figure}

\begin{figure}[!htb]
  \centering
  \includegraphics[width=.9\textwidth]{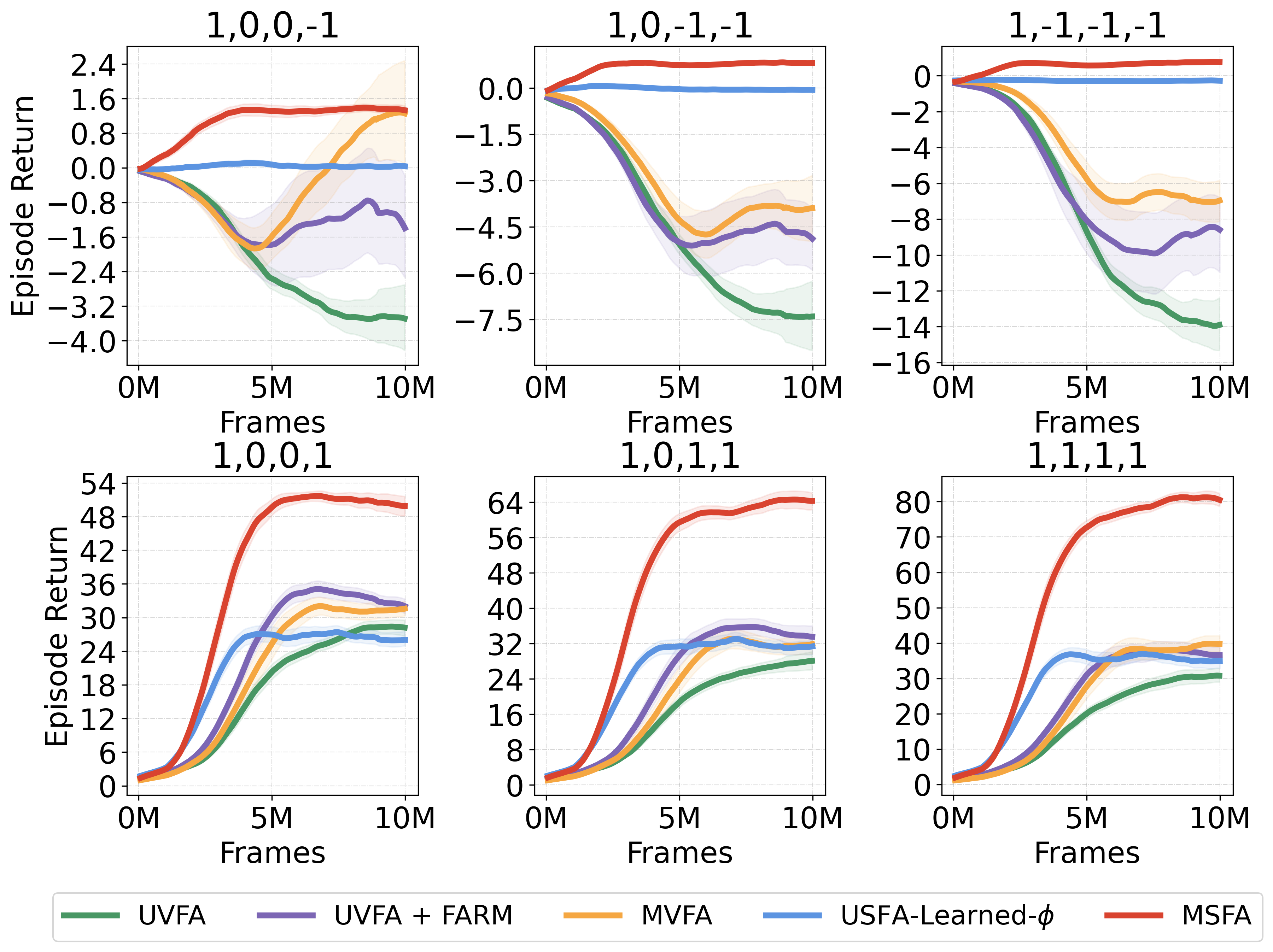}
  \caption{All results for the Fruitbot environment. When objects must be avoided, we see that no method does well, though MSFA generalizes best and can pick up a couple of objects. When combinations of objects must be collected, MSFA outperforms baselines by a large margin.
  }
\end{figure}

\clearpage

\section{Hyperparameters}\label{appendix:hps}
We shown hyperparameters in Tables \ref{tab:shared} and \ref{tab:algos}. Tables \ref{tab:shared} describes hyperparameters that were shared across algorithms and Tables \ref{tab:algos}  describes hyperparameters that were different across algorithms. We developed the algorithms using the \citet{hoffman2020acme} reinforcement learning codebase and unless stated otherwise using default ``R2D2''~\citep{kapturowski2018recurrent} config values in the codebase. We describe our development and hyperparameter search in more detail below.

\subsection{Development and hyperparameter search}

The first thing we did was replicate the generalization gap between USFA and UVFA that~\citet{borsa2018universal} produced in their paper using the BabyAI gridworld~\citep{chevalier2018babyai}. As a result, most hyperparameters were tuned for either USFA or UVFA. We used the Atari Convolutional network used by ``Deep Q-Networks'' (DQN) in~\citet{mnih2015human} as our visual encoder and an LSTM~\citep{hochreiter1997long} as our state representation function.

Once we were able to replicate their generalization performance, we implemented USFA-Learned-$\phi$ and MSFA. In this setting, we are learning Q-values, successor features, and predicting rewards from cumulants. This leads to the following coefficients for losses: $\alpha_Q$, $\alpha_\psi$, and  $\alpha_\phi$ respectivey. We fixed $\alpha_\psi=1$ and searched $\alpha_\phi \in [.01, .1, 1, 10, 100]$ and  $\alpha_Q \in [.01, .05, .1,  .5, 1.0]$ for both USFA-Learned-$\phi$ and MSFA. For MSFA and UVFA-FARM, we chose the number of modules to be the number of training tasks. We set each module size so that the number of parameters by all algorithms was approximately the same (with USFA as a reference). We found that FARM hyperparameter~\citep{carvalho2021feature} worked well for both UVFA-FARM and for MSFA (FARM modules are the basis of MSFA state-modules). We highlight that properly masking all losses was absolutely critical to good generalization performance. Without proper masking, out-of-episode data is used for value estimation of in-episode data, which can lead to inaccurate value estimates. We suspect that this hampers GPI. 

When doing experiments for Fruitbot, we took all implementation details from~\citep{cobbe2020leveraging}. We first replicated the DQN performance from~\citet{cobbe2020leveraging} with UVFA. Some important changes that we made included (1) adding prioritized experience replay (2) leveraging the Impala ResNet vision torso~\citep{espeholt2018impala} (3) lowering the memory size to match the paper (4) using the ${\tt SIGNED\_HYPERBOLIC\_PAIR}$ value transformation for UVFA-based algorithms (5) increasing the capacity of the Q-value estimation MLPs. We applied these settings to other baselines and redid our search over $\alpha_\phi$ and $\alpha_Q$ for both MSFA and USFA-Learned-$\phi$. For each method, we changed the size of the networks so they were approximately the same (with UVFA as a reference). Since MSFA uses disentangled functions for $\phi$ and $\psi$, it has fewer parameters.

When doing experiments in Minihack, we found that the Fruitbot changes were important for good performance in this domain. All hyperparameters led to strong training performance but poor generalization for UVFA. We found that increasing the batch size was important for improving generalization of USFA-Learned-$\phi$ and MSFA but was detrimental to UVFA-based methods. 

\begin{table}[!htp]\centering
  \caption{Hyperparameters shared across all algorithms.}\label{tab:shared}
  \scriptsize
  \begin{tabular}{lrrrr}\toprule
  \textbf{Algorithm} &\textbf{} &\textbf{} &\textbf{} \\\midrule
  \textbf{Loss Hyperparameters} &BabyAI &Fruibot &Minihack \\\midrule
  discount &0.99 & 0.99& 0.99\\
  burn\_in\_length &0 &0 &0 \\
  trace\_length &40 &40 &40 \\
  importance\_sampling\_exponent &0 &0.6 &0.6 \\
  max replay size &100,000 &70,000 &70,000 \\
  max gradient norm &80 &80 &80 \\\midrule
  \textbf{Shared Network Components} & & & \\\midrule
  Vision torso &DQN ConvNet &Impala ResNet &Impala ResNet \\
  \bottomrule
  \end{tabular}
  \end{table}

\begin{table}[!htp]\centering
  \caption{Hyperparameters for individual algorithms. Note: $T_1={\tt IDENTITY\_PAIR}$ and $T_2={\tt SIGNED\_HYPERBOLIC\_PAIR}$. }\label{tab:algos}
  \scriptsize
  \begin{tabular}{lrrrr}\toprule
   &BabyAI &Fruibot &Minihack \\\midrule
  \textbf{MSFA} & & & \\\midrule
  Parameters (millions) & 2.1M & 1.66M & 1.7M \\
  $\alpha_\phi$ &1 &1 &1 \\
  $\alpha_\psi$ &1 &1 &1 \\
  $\alpha_Q$ &0.5 &0.5 &0.5 \\
  module\_size &150 &60 &80 \\
  nmodules &4 &4 &3 \\
  attention heads &2 &2 &1 \\
  batch size &32 &32 &64 \\
  $\phi$ MLP hidden sizes &[256] &[256] &[256] \\
  $\psi$ MLP hidden sizes &[128] &[512, 512] &[512, 512] \\
  projection dim &16 &16 &16 \\
  \midrule
  \textbf{USFA} & & & \\\midrule
  Parameters (millions) & 2.35M & 1.98M & 1.95M \\
  lstm size &512 &300 &300 \\
  $\alpha_\phi$ &1 &1 &1 \\
  $\alpha_\psi$ &1 &1 &1 \\
  $\alpha_Q$ &0.5 &0.5 &0.5 \\
  batch size &32 &32 &128 \\
  $\phi$ MLP hidden sizes &[256] &[256] &[256] \\
  $\psi$ MLP hidden sizes &[128] &[512, 512] &[512, 512] \\
  \midrule
  \textbf{UVFA} & & & \\\midrule
  Parameters (millions) & 2.09M & 1.96M & 1.95M \\
  lstm size &512 & 256 & 256 \\
  rlax.TxPair & $T_1$ & $T_2$ &$T_2$ \\
  batch size &32 &32 &32 \\
  Q MLP hidden sizes &[128] &[512, 512] &[512, 512] \\
  \midrule
  \textbf{UVFA-FARM} & & & \\\midrule
  Parameters (millions) & 2.17M & 2.15M & 2.06M \\
  module\_size &150 &64 &80 \\
  nmodules &4 &4 &3 \\
  attention heads &2 &2 &1 \\
  rlax.TxPair & $T_1$ & $T_2$ &$T_2$ \\
  batch size &32 &32 &32 \\
  projection dim &16 &16 &16 \\
  Q MLP hidden sizes &[128] &[512, 512] &[512, 512] \\
  \bottomrule
  \end{tabular}
  \end{table}

\section{Environments} \label{appendix:envs}
We presented most environment information in the main text. Here, we specify which levels we used from the corresponding environments or how the changed environments for our experiments.

\subsection{Procgen}
We used the Fruitbot environment in Procgen. We made the following changes. We set the maximum number of levels that an agent could complete within a lifetime to $4$. We divided the 12 object types $\{O_1 \ldots O_{12}\}$ into 4 categories: $C_1=\{O_1, O_{2}, O_{3}\}, \ldots, C_4=\{O_{10}, O_{11}, O_{12}\}$, where each category provided postive reward for picking up any of its 3 category types.

\subsection{Minihack}
We did not make any changes to the environment. Train tasks  were: ``MiniHack-Room-Monster-Y-v0'', ``MiniHack-Room-Trap-Y-v0'', and ``MiniHack-Room-Dark-Y-v0''. Test tasks  were ``MiniHack-Room-Ultimate-Y-v0''. We used $Y=5 \times 5$ for the ``small room'' experiments and $Y=15 \times 15$ for the ``large room'' experiments.

\end{document}